\DocumentMetadata{%
	pdfstandard=A-3u,
	pdfversion=1.7,
	lang=en-US,
}

\documentclass[nofoot, balance,colorlinks,upint,subscriptcorrection,varvw,mathalfa=cal=boondoxo,spanish,german,vietnamese,russian,greek]{asmeconf}

\hypersetup{%
	pdfauthor={John H. Lienhard},									  
	pdftitle={ASME Conference Paper LaTeX Template},                  
	pdfkeywords={ASME conference paper, LaTeX template, BibTeX style},
	pdfsubject = {Describes the asmeconf LaTeX template},			  
	pdflicenseurl={https://ctan.org/pkg/asmeconf},
}

\usepackage{amsmath}
\usepackage{subcaption}
\usepackage{algorithm}
\usepackage{algorithmic}

\begin{document}



\title{A Domain Adaptation of Large Language Models for Classifying Mechanical Assembly Components} 
 
%
%
%

\SetAuthors{%
    Fatemeh Elhambakhsh\affil{1}, 
	Daniele Grandi\affil{2},  
	Hyunwoong Ko\affil{3}\CorrespondingAuthor{Hyunwoong.Ko@asu.edu}
	}

\SetAffiliation{1}{School of Computing and Augmented Intelligence, Arizona State University, Tempe, AZ}
\SetAffiliation{2}{Autodesk, San Francisco, CA}
\SetAffiliation{3}{School of Manufacturing Systems and Networks, Arizona State University, Mesa, AZ}


\maketitle





\begin{abstract}
The conceptual design phase represents a critical early stage in the product development process, where designers generate potential solutions that meet predefined design specifications based on functional requirements. Functional modeling, a foundational aspect of this phase, enables designers to reason about product functions before specific structural details are determined. A widely adopted approach to functional modeling is the Function-Behavior-Structure (FBS) framework, which supports the transformation of functional intent into behavioral and structural descriptions. However, the effectiveness of function-based design is often hindered by the lack of well-structured and comprehensive functional data. This scarcity can negatively impact early design decision-making and hinder the development of accurate behavioral models. Recent advances in Large Language Models (LLMs), such as those based on GPT architectures, offer a promising avenue to address this gap. LLMs have demonstrated significant capabilities in language understanding and natural language processing (NLP), making them suitable for automated classification tasks. This study proposes a novel LLM-based domain adaptation (DA) framework using fine-tuning for the automated classification of mechanical assembly parts’ functions. By fine-tuning LLMs on domain-specific datasets, the traditionally manual and subjective process of function annotation can be improved in both accuracy and consistency. A case study demonstrates fine-tuning GPT-3.5 Turbo on data from the Oregon State Design Repository (OSDR), and evaluation on the A Big CAD (ABC) dataset shows that the domain-adapted LLM can generate high-quality functional data, enhancing the semantic representation of mechanical parts and supporting more effective design exploration in early-phase engineering.
\end{abstract}



\section{Introduction}
The process of translating ideas to developing new products or enhancing an existing product to meet specific objectives that can address various aspects, such as usability, functionality, sustainability, and aesthetics, is known as product design \cite{baxter2018product}. 
Thanks to computer-aided product tools, such as computer-aided design (CAD) and computer-aided manufacturing (CAM), product design processes have become more digitalized
\cite{zhang2020digital}. The product design phase consists of three main stages: (1) Specification, (2) Conceptual design, and (3) Detailed design \cite{ibbotson2018approach}. The conceptual design stage is arguably the critical stage, for it involves functional formulation and concept generation, evaluation, and improvement to generate physical solutions that meet design specifications \cite{wang2020digital, hsu1998current}. One way to achieve this goal is by utilizing the Function-Behavior-Structure (FBS) framework \cite{al2016function}.

The FBS framework refers to conceptualizing objects based on their function, structure, and behavior, which mainly transforms presumed functions into behavioral descriptions \cite{al2016function}. 
A function is defined as a set of tasks that enable a product to fulfill its design purpose \cite{ferrero2022classifying}. Functional modeling supports and offers decision guidance for a product design to understand an overall product function during the conceptual design phase, where design decisions are yet to be made \cite{ferrero2022classifying, hirtz2002functional}.
Although systematic, function-based design can be subjective due to how designers interpret the intended function. The human interpretation of function has often generated unstructured and unbalanced repositories, thereby affecting the accuracy of behavior predictions and subsequent structural decisions. Therefore, a significant challenge arises from the limited availability of well-structured function data, necessitating a more accurate, automated way of generating function data. 

With the advent of large language models (LLMs), such as GPTs and Llama, language understanding and generation have been significantly advanced by training billions of parameters on vast amounts of text \cite{raiaan2024review}.
LLMs can identify patterns and relationships within language and apply these to generating responses, solving problems, and writing. LLMs have the ability to perform a wide range of natural language processing tasks without the need for task-specific training because they learn broader language patterns and relationships rather than being confined to specific tasks \cite{ray2023chatgpt}.
This versatility makes LLMs ideally suited for various language-related tasks.

However, directly applying LLMs to complex domain-specific contextual applications faces many challenges caused by heterogeneity, uniqueness, sophistication, and constraints of domain data \cite{ling2023domain}. Consequently, efforts have shifted toward adapting foundational, pre-trained, general-purpose LLMs to domain-specific applications. Design and manufacturing are particularly critical application areas, as the knowledge within these fields contains complex structures due to the intricate nature of a product manufacturing lifecycle from design to build. This necessitates domain adaptation (DA) to bridge the knowledge gap between foundational LLM capabilities and the distinct characteristics of design and manufacturing datasets.

Given the capabilities of LLMs in learning and generating textual data,
as well as the challenge of the lack of functional data in mechanical design,
this study explores two research questions: (1) Are LLMs sufficiently capable
of generating accurate functional annotations for mechanical components? 2)
How effectively can customizing LLMs infer the functions of new mechanical
components based on textual data? To address the research questions, the study
proposes an LLM DA framework for the function classification of mechanical assembly
parts. The case study utilizes the Oregon State Design Repository (OSDR) as the domain knowledge LLM customization. Later, the study applied the domain-adapted LLM to A Big
Cad (ABC) data set to generate novel classified functional data.

The remainder of the paper is as follows: Section 2 reviews the literature.
Section 3 introduces the problem statement. Section 4 proposes the methodology. Section 5 presents a case study. Section 6 discusses the advantages and
limitations. Section 7 concludes the study.

\section{Literature Review}
This section reviews the latest studies of functional modeling and the role of LLMs in mechanical design.
\subsection{Function Annotation in Mechanical Design}
Function refers to a system's general input and output relationship that performs a task \cite{kirschman1998classifying}. In mechanical design, function refers to the set of tasks that enable a product to perform its intended function \cite{ferrero2022classifying}. Function-based tools and modeling offer decision guidance for product design to understand an overall product function during the conceptual design phase, where design decisions are yet to be made \cite{ferrero2022classifying, hirtz2002functional}. 

In this regard, Ferrero et al. proposed a graph neural network (GNN)-based framework to classify mechanical component functions to address the scarcity of function-based design data \cite{ferrero2022classifying}. Han et al. 
\cite{han2023automatic} developed a three-dimensional (3D) product-based methodology to construct a semantic functional ontology for semantic annotation of functions for mechanical parts in conceptual design. The goal was to bridge the gap to capture the correlation of function and structure and retrieve 3D models based on their specific functions at the early design stage.

\subsection{Large Language Models in Mechanical Design}

Language is the way humans communicate, self-express, and interact with machines \cite{naveed2023comprehensive}. The need for language-based models arises from the challenges of handling complex language-based tasks, such as translation, summarization, information retrieval, and conversations with the machines \cite{naveed2023comprehensive}.
Recent advances in Natural Language Processing (NLP), driven by transformers, enabled the development of Large Language Models (LLMs) that achieve near-human performance across various domains \cite{chernyavskiy2021transformers}.

LLMs demonstrate remarkable capabilities and advantages in performing complex NLP tasks, such as capturing the underlying data distribution of sequential, contextual data and generating content that closely mimics the patterns of their training data \cite{hadi2023survey, farahmand2025attengluco, alsadat2024using}. However, directly applying LLMs to complex domain-specific contextual applications faces many challenges caused by heterogeneity, uniqueness, sophistication, and constraints of domain data \cite{ling2023domain}. Consequently, efforts have shifted toward adapting foundational, pre-trained, general-purpose LLMs to domain-specific applications. In general, there are three approaches for LLM DA in literature : (1) External augmentation, (2) Prompt crafting, and (3) Model fine-tuning, utilizing them heavily depends on the level of accessibility of LLMs \cite{ling2023domain}.

Several studies utilized the capabilities of LLMs in conceptual design. For instance, Chen et al. proposed an LLM-guided framework to generate design concepts using the FBS method \cite{chen2024toward}. Li et al. developed a method, LLM4CAD, to
enable multimodal LLMs in 3D CAD generation utilizing different input modalities,
including the text-only, text+sketch, text+image, and text+
sketch+image data \cite{li2024llm4cad}. Ataei et al. developed an LLM-based framework, Elicitron, that utilizes
to automate and improve requirements elicitation in product
design \cite{ataei2025elicitron}. The study utilized LLM agents to simulate a wide range of
user personas, which can explore diverse and latent needs that
traditional methods overlooked. 

While these studies showcase the promising role of LLMs in conceptual design, they still face limitations, such as not fully utilizing domain-specific knowledge through DA techniques.  Design and manufacturing are particularly
critical application areas, as the knowledge within these fields
contains complex structures due to the intricate nature of a product manufacturing lifecycle from design to build. This highlights the need for LLM customization techniques to fully capture domain-specific knowledge. 
\section{PROBLEM STATEMENT}
Functional modeling supports and offers decision guidance for a product design to understand an overall product function during the conceptual design phase, where design decisions are yet to be made \cite{ferrero2022classifying, hirtz2002functional}. However, function-based design is subjective due to how designers interpret the intended function. The human interpretation of function has often generated unstructured and unbalanced repositories, thereby affecting the accuracy of behavior predictions and subsequent structural decisions. Therefore, a significant challenge arises from the limited availability of well-structured function data, necessitating a more accurate, automated way of generating function data. 

The traditional ways of function annotations for mechanical assembly components are mainly based on manually curated data or domain-specific rules. These methods are not only time-consuming and inaccurate, but they cannot capture the complex correlation between structure and function. These challenges limit traditional methods' scalability and adaptability for conceptual design generation. Recent studies have explored the capabilities of Machine Learning (ML), such as GNNs, to automatically classify mechanical parts' function \cite{ferrero2022classifying}. Although graph models have demonstrated strong
capabilities across various domains \cite{sabri2020monitoring, elhambakhsh2021developing, elhambakhsh2022scan, elhambakhsh2023latent}, these methods lack the mechanism to capture the complex distribution of sequential, contextual data.

To address the challenge of lack of function data and the limited capability of traditional ML methods, this study proposed a supervised DA method for LLMs to generate synthesized function-classified data \cite{parthasarathy2024ultimate}.

\section{METHODOLOGY}
\begin{figure*}[tbh]
    \centering
    \includegraphics[width=\textwidth]{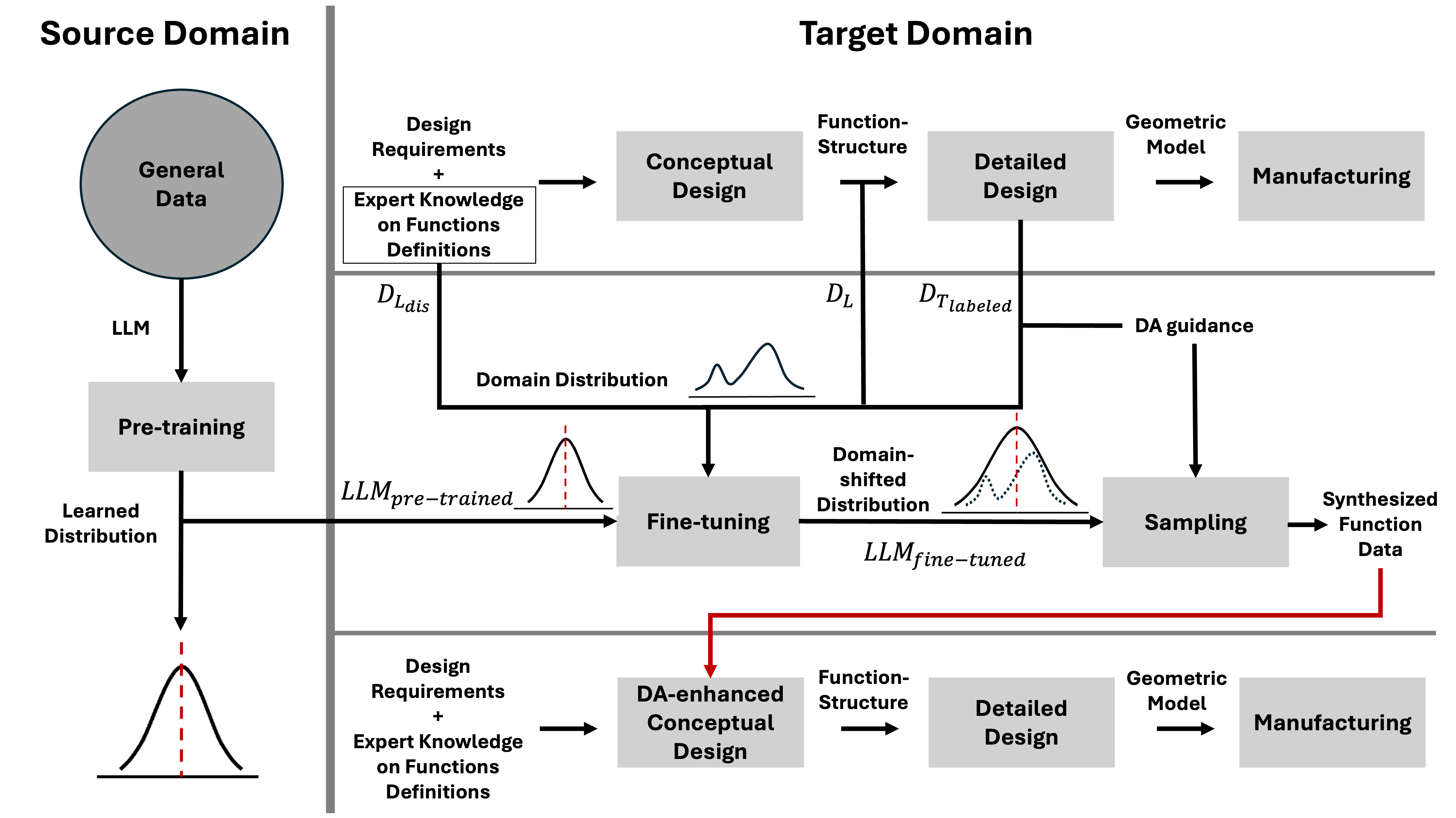}
    \caption{An overview of a supervised DA via fine-tuning}
    \label{DA}
\end{figure*}

This section introduces a supervised DA framework utilizing a fine-tuning approach. The method aims to automatically classify and annotate mechanical assembly components to generate synthesized functional labels in the conceptual design of a product design. Figure \ref{DA} indicates the overview of the proposed framework.

DA relies on two distinct domains: a source environment (general domain) and a target domain (assembly design domain). In this study, a generative language model, pre-trained on the source domain to capture the general population distribution of the source data, adapts its knowledge of assembly design labels from the source domain to improve performance in the target domain \cite{goodman2023supervised}. The goal is to fine-tune the model’s distribution toward assembly data. 

To achieve this, the study employs function-labeled contextual design data as the target domain for supervised fine-tuning, serving as the DA method. 
The function labels are derived from the training dataset’s conceptual design phases, while the assembly data represent the mechanical parts and their assembly name of the outputs of the corresponding detailed design phases. Once fine-tuned, the language model generates synthesized label samples of assembly designs based on detailed design phases for \textit{the target domain's different design cases/scenarios}. These synthesized samples, guided by the fine-tuned distribution, enhance the function-structure database used in the target domain's design processes, ultimately supporting the transition from conceptual to detailed design phases.


Function labels play a crucial role in guiding the generative model’s learning process by mapping design features to their intended functionalities. These labels categorize design elements based on their purpose, enabling the model to generate meaningful assembly design samples aligned with conceptual design requirements. For example, in an assembly design domain, function labels may include “support,” “fasten,” “transmit motion,” or “seal,” each defining a specific role in the overall design structure.

The function-labeled contextual design data (\(D_{T_{\text{labeled}}}\)) indicates the relationships between design components and their intended functionalities. \(D_{T_{\text{labeled}}}\) consists of \(n\) columns, where the first \(n-1\) columns represent design features, and the final column contains function labels (classes) that guide the training process by demonstrating the desired behavior for the model. 
This structured data format is designed to align with the model’s learning process, where separating features from function labels allows the transformer-based architecture to establish correlations between design characteristics and functional intent. By structuring the dataset in this way, the model can effectively capture latent relationships between design features and their corresponding functionalities, improving generalization and predictive accuracy in generating assembly designs.

The function labels data (\(D_L\)) comprises a set of \(m\) function classes that define the labels of interest in the conceptual design phase for the training process.
Additionally, the function definition dataset (\(D_{L_{\text{dis}}}\)) provides textual descriptions or structured representations of these \(m\) function classes, serving as a reference for interpreting and validating the assigned labels\cite{stone1999development}. Unlike traditional design methods, which rely on manual classification and domain expertise to assign functions, the incorporation of textual descriptions uniquely enhances labeling and design processes through the transformer model’s ability to understand contextual information \cite{vaswani2017attention}. 

The self-attention mechanism in transformers enables the model to extract nuanced relationships between words and concepts, allowing it to recognize functional similarities even when design elements vary in structure or composition. This capability empowers the model to automate and refine the labeling process, reducing human effort and improving consistency in function classification. By leveraging structured function labels and contextual descriptions, the model learns a more comprehensive representation of assembly design principles, ultimately enhancing its ability to generate functionally coherent synthesized label data that support the conceptual design phase.


We defined three hyper-parameters, $H = [E, B, LR]$, where $E$, $B$, and $LR$, are the number of epochs, batch size, and learning rate multiplier, respectively, to enhance LLM's performance. During each tuning iteration, we randomly select and evaluate combinations of hyperparameters
from a specified range to explore a large parameter space as written in Equation \ref{RS}:

\begin{equation} \label{RS}
H = \{(E, B, LR) \mid E \in \mathcal{E}, B \in \mathcal{B}, LR \in \mathcal{L} \}
\end{equation}

The next step is the fine-tuning step, which is the main step of the proposed pipeline. 
Utilizing the randomly chosen hyperparameters, the LLM receives $D_{T_{\text{labeled}}}$, $D_L$, $D_{L_{\text{dis}}}$ as the input with a prompt and predicts the function labels from $D_L$. The fine tuning process is represented in Equation \ref{finetuned} \cite{li2025integrating}:
\begin{equation}\label{finetuned}
LLM_{\text{pre-trained}}(\{D_{T_{\text{labeled}}}, D_L, D_{L_{\text{dis}}}\}) \rightarrow LLM_{\text{fine-tuned}}
\end{equation}

More specifically,  the LLM utilizes the prompt to receive the tokenized input ($d_1,...,d_s$) and predict label y from $D_L$.
Equation \ref{soft} indicates the output for the final layer of the transformer in the LLM which predicts $y$:

\begin{equation} \label{soft}
P(y \mid d_1, \dots, d_s) = \text{softmax}(h_s^l W_y)
\end{equation}
, where $h_s^l$ and $W_y$ are the hidden state for the final layer and weight matrix for the classification task, respectively.
During each fine-tuning iteration, 
The objective function, which is based on the minimization of the cross entropy function is shown in Equation \ref{cross}:
\begin{equation}\label{cross}
L(D_{T_{\text{labeled}}}, D_L, D_{L_{\text{dis}}}) = - \sum_{(d, y)} \log P(y \mid d_1, \dots, d_s)
\end{equation}

The final objective function and updating hyperparameters are shown in Equations \ref{obj}, and \ref{w}:
\begin{equation}\label{obj}
    L(D_{T_{\text{labeled}}}, D_L, D_{L_{\text{dis}}})  = - \sum_{e=1}^{E} \sum_{i=1}^{\frac{N}{B}} \sum_{(d, y) \in \mathcal{B}_i} \log P(y \mid d_1, \dots, d_s)
\end{equation}

\begin{equation}\label{w}
W_y^{t+1} = W_y^t - LR \cdot \nabla_{W_y} L(D_{T_{\text{labeled}}}, D_L, D_{L_{\text{dis}}})
\end{equation}
, where $N$ is the total numebr of samples, $i$ is the i-th batch, and 
\\$\nabla_{W_y} L(D_{T_{\text{labeled}}}, D_L, D_{L_{\text{dis}}})$ is the gradient of the loss function.

We applied the fine-tuned LLM to a test function labeled mechanical parts data ($D_{\text{test}}$) to predict the labels and evaluate the performance of the LLM during the evaluation step.
For evaluation, we utilized accuracy, precision, recall, and F1 metrics as shown in Equation \ref{Ini}:

\begin{equation}\label{Ini}
M = \frac{1}{S} \sum_{i=1}^{S} \mathcal{F}(\hat{y}_i, y_i)
\end{equation}
, where $M$ is an evalution metric, $S$ indicates the number of samples for $D_{T_{\text{labeled}}}$, $\mathcal{F}$ is the evaluation function, and $\hat{y}$ is the predicted function class for each mechanical part.

The evaluation step is necessary to choose a list of optimized hyperparameters as in Equation \ref{opt}:
\begin{equation} \label{opt}
{H}' = \{(E', B', LR') \mid E' \in \mathcal{E}, \, B' \in \mathcal{B}, \, LR' \in \mathcal{L}\}
\end{equation}
We selected the 
fine-tuned LLM with the highest values of evaluation metrics to choose ${H}'$. Algorithm 1 indicates the fine-tuning steps.

Finally, the design domain-adapted classifier with the optimized hyperparameters values is applied for the function annotation of unlabeled design data ($D_{\text{unlabeled}}$) for a different design scenario. The classification task is represented in Equation \ref{class}:

\begin{equation}\label{class}
    LLM_{\text{fine-tuned}}(D_{\text{unlabeled}}, D_L, D_{L_{\text{dis}}}) \rightarrow \hat{y}(D_{\text{unlabeled}})
\end{equation}
, where $\hat{y} \in D_L$ is the predicted function for a mechanical part in the unlabeled design data. The study applied the proposed domain-adapted LLM classifier to a conceptual design case study to generate synthesized function annotations for mechanical assembly components.

\begin{algorithm}[H]
\caption{SUPERVISED DA VIA FINE-TUNING}
\label{alg:fine_tuning}
\begin{algorithmic}[1]
    \STATE \textbf{INPUT:}  $\text{LLM}_{\text{pre-trained}}$,  $D_{T_{labeled}}$, $D_L$, $D_{L_{dis}}$, and $H=[E, B, LR]$
    \STATE \textbf{OUTPUT:} ${H}' = [E', B', LR']$, and $\text{LLM}_{\text{fine-tuned}}$

    \STATE Initialize hyperparameter combinations: $H = \{(E,B,LR) | E \in \mathcal{E}, B \in \mathcal{B}, LR \in \mathcal{LR}\}$

    \FOR {each hyperparameter set $\left(E,B,LR\right) \in H$}
        \STATE Initialize model: $\text{LLM}_{\text{fine-tuned}} \leftarrow \text{LLM}_{\text{pre-trained}}$ 
        \FOR {epoch $e = 1,2,\ldots,E$}
            \FOR {each batch $\left(d,y\right) \in D_{T_{labeled}}$}
                \STATE Compute loss using Equation (5) 
                \STATE Compute gradient using Equation (6)
                \STATE Update parameters using Equation (6)
                \STATE Compute predictions using Equation (3)
            \ENDFOR
        \ENDFOR
        \STATE Evaluate model using Equation (7)
    \ENDFOR

    \STATE Select optimal hyperparameters $\left(E', B', LR'\right)$ using Equation (8)

    \RETURN ${H}' = [E', B', LR']$, and $\text{LLM}_{\text{fine-tuned}}$
\end{algorithmic}
\end{algorithm}

\section{CASE STUDY}
\begin{figure*}[tbh]
    \centering
    \includegraphics[width=\textwidth, height=\textheight, keepaspectratio]{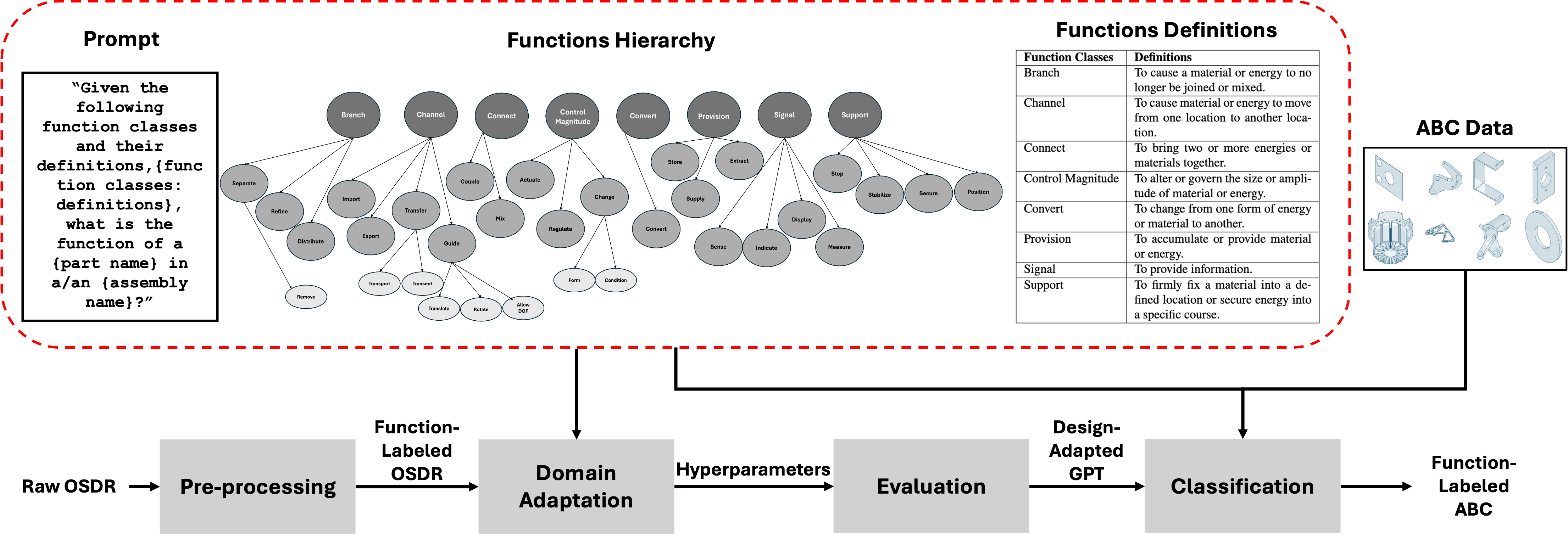}
    \caption{Function Classification of Mechanical Assembly Parts}
    \label{CASE}
\end{figure*}

The case study section indicates the capability of the proposed method for a function annotation task for mechanical assembly parts. The section, initially, explains the datasets used for DA and classification steps. Then, the experimental details of fine-tuning are highlighted. Finally, we discuss the results. Figure \ref{CASE} provides an overview of the case study.

\subsection{Data}
\subsubsection{Oregon State Design Repository}

\begin{figure*}[h!]
    \centering
    \includegraphics[width=1\textwidth, height=0.3\textheight]{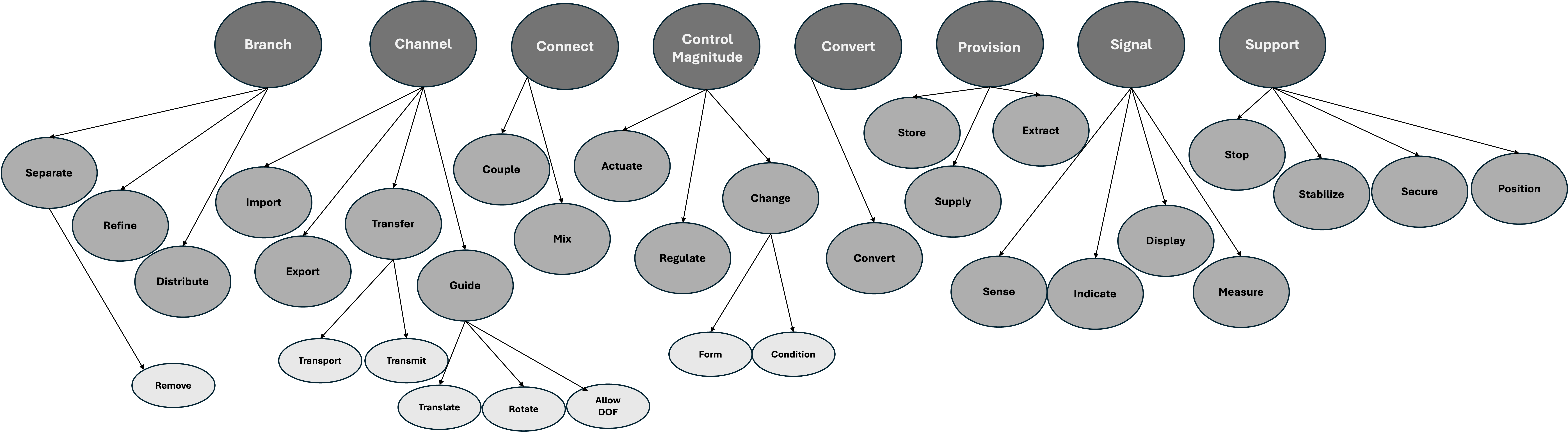}
    \caption{A three-tier function hierarchy \cite{stone1999development}}
    \label{Function}
\end{figure*}

The OSDR includes information on various mechanical assembly parts, their basis, system type, function, sub-function, flow, and assembly relationships \cite{bohm2004product, bohm2008introduction}. 
The OSDR contains 15636 samples. We removed the duplicate product values and the ones with a lack of information regarding their basis, system name, and function. The pre-processed OSDR contains 7568 unique parts. Table \ref{tab:osdr_final} is an illustration of the pre-processed OSDR dataset. The OSDR includes the products' functions and sub-functions in a three-tier hierarchy. We extracted the function hierarchy from the data as shown in Figure \ref{Function}. The first-tier functions are the labels of interest in this study for the classification task, containing eight function classes. Table  \ref{tab:function_classes} indicates the definition of function classes \cite{stone1999development}.

\begin{table*}[h!]
\centering
\caption{Pre-processed Function-labeled Design Data}
\renewcommand{\arraystretch}{0.8}  
\begin{tiny}  
\resizebox{\textwidth}{!}{%
\begin{tabular}{|l|l|l|l|l|}
\hline
\textbf{name} & \textbf{component\_basis} & \textbf{sys\_name} & \textbf{subfunction\_basis} & \textbf{parent\_subfunction} \\ \hline
master cylinder & hydraulic pump & brake system & import & channel \\ \hline
drum            & container      & brake system & import & channel \\ \hline
caliper         & clamp          & brake system & import & channel \\ \hline
caliper         & clamp          & brake system & export & channel \\ \hline
caliper         & clamp          & brake system & regulate & control magnitude \\ \hline
pedal           & lever          & brake system & guide & channel \\ \hline
fin             & airfoil        & brake system & distribute & branch \\ \hline
fin             & airfoil        & brake system & export & channel \\ \hline
brake cylinder  & hydraulic pump & brake system & export & channel \\ \hline
rotor           & wheel          & brake system & import & channel \\ \hline
battery tray    & housing        & digger dog   & transfer & channel \\ \hline
\end{tabular}%
}
\end{tiny}
\label{tab:osdr_final}
\end{table*}

\begin{table}[tbh]
\centering
\caption{Function Classes and their Definitions \cite{stone1999development}}
\resizebox{\columnwidth}{!}{%
\begin{tabular}{|l|p{5cm}|}
\hline
\textbf{Function Classes} & \textbf{Definitions} \\ \hline
Branch & To cause a material or energy to no longer be joined or mixed. \\ \hline
Channel & To cause material or energy to move from one location to another location. \\ \hline
Connect & To bring two or more energies or materials together. \\ \hline
Control Magnitude & To alter or govern the size or amplitude of material or energy. \\ \hline
Convert & To change from one form of energy or material to another. \\ \hline
Provision & To accumulate or provide material or energy. \\ \hline
Signal & To provide information. \\ \hline
Support & To firmly fix a material into a defined location or secure energy into a specific course. \\ \hline
\end{tabular}%
}

\label{tab:function_classes}
\end{table}

\subsubsection{Geometric Model Data}
\begin{figure}[h!]
    \centering
    \includegraphics[width=\columnwidth]{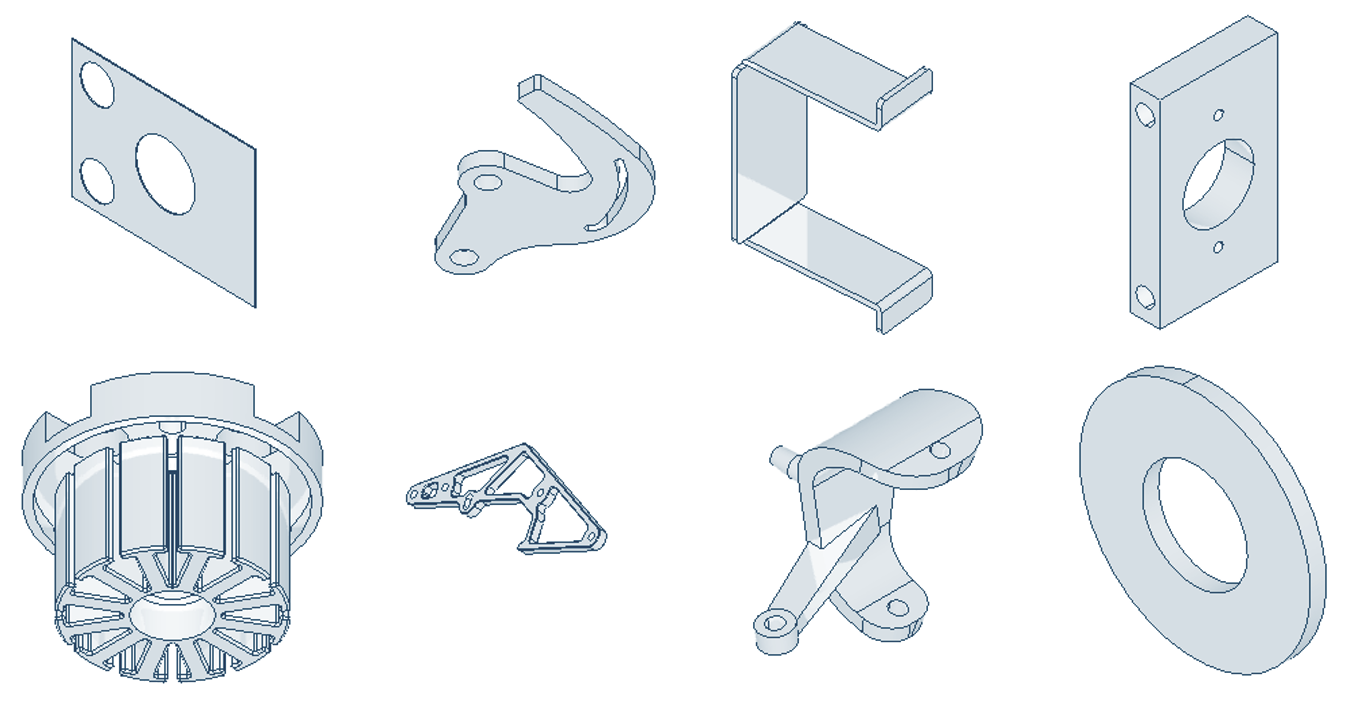}
    \caption{SAMPLES OF MECHANICAL PARTS IN ABC DATA \cite{koch2019abc}.}
    \label{ABC}
\end{figure}

The ABC data set consists of one million high-quality CAD geometric models of mechanical parts \cite{koch2019abc}. The models are defined by parametric surfaces and are accompanied by accurate ground truth information regarding patch decomposition, sharp feature annotations, and differential properties. Figure \ref{ABC} indicates samples of CAD models for mechanical parts in the ABC data. The dataset is split into 100 chunks, each of which contains 10000 CAD models of mechanical parts. We matched the samples' names in each chunk with those in the OSDR dataset. 
The matched dataset contains 27950 matched samples. We removed the duplicate samples. Finally, the matched dataset contains 6786 unique ABC mechanical parts with their names and their assembly names.

\subsection{Fine-Tuning}
\begin{figure*}[h!]
  \centering
  \begin{subfigure}[b]{0.45\textwidth}
    \centering
    \includegraphics[width=\textwidth]{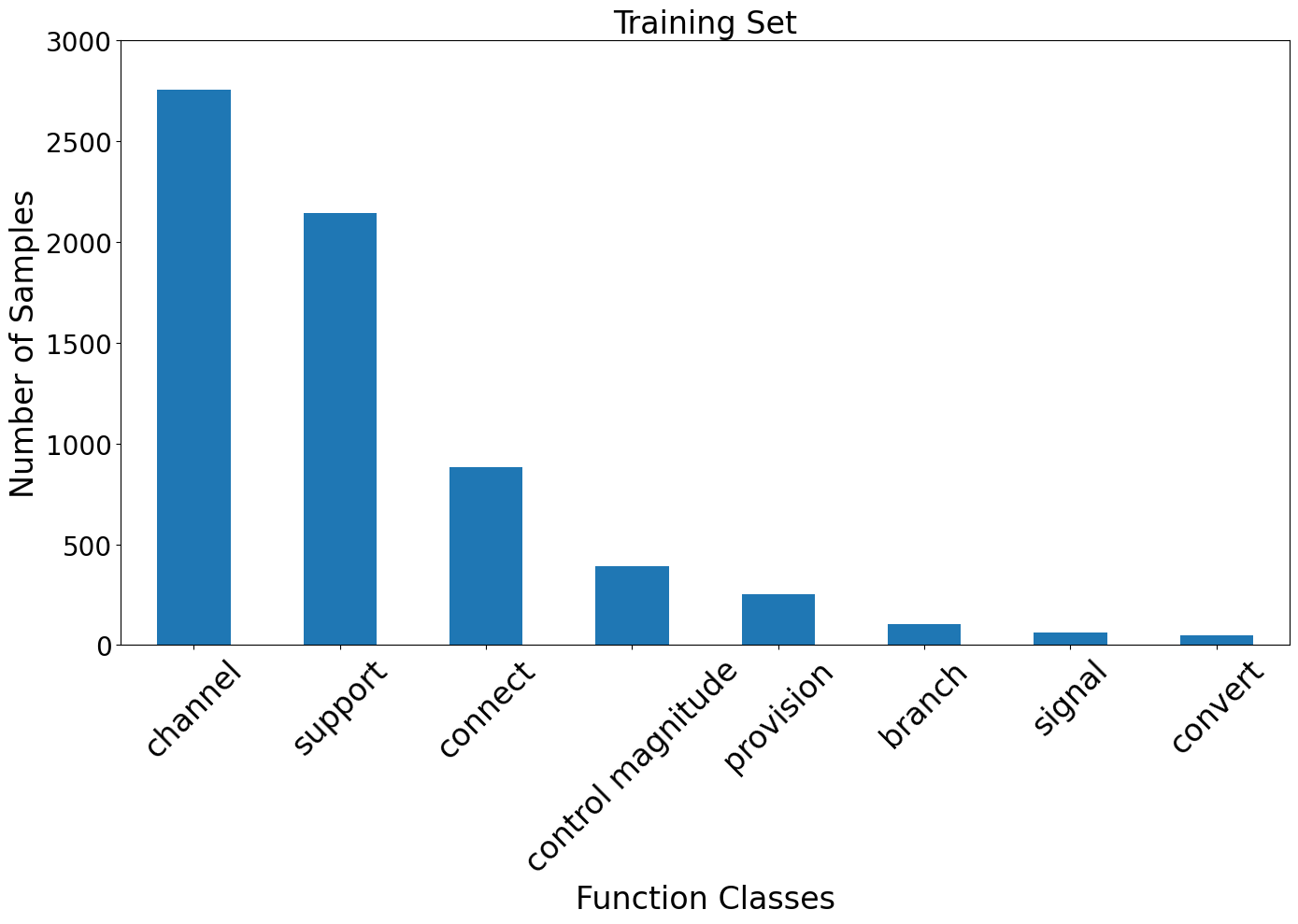}
    \caption{Train set.}
    \label{fig:sub1}
  \end{subfigure}
  \hfill
  \begin{subfigure}[b]{0.45\textwidth}
    \centering
    \includegraphics[width=\textwidth]{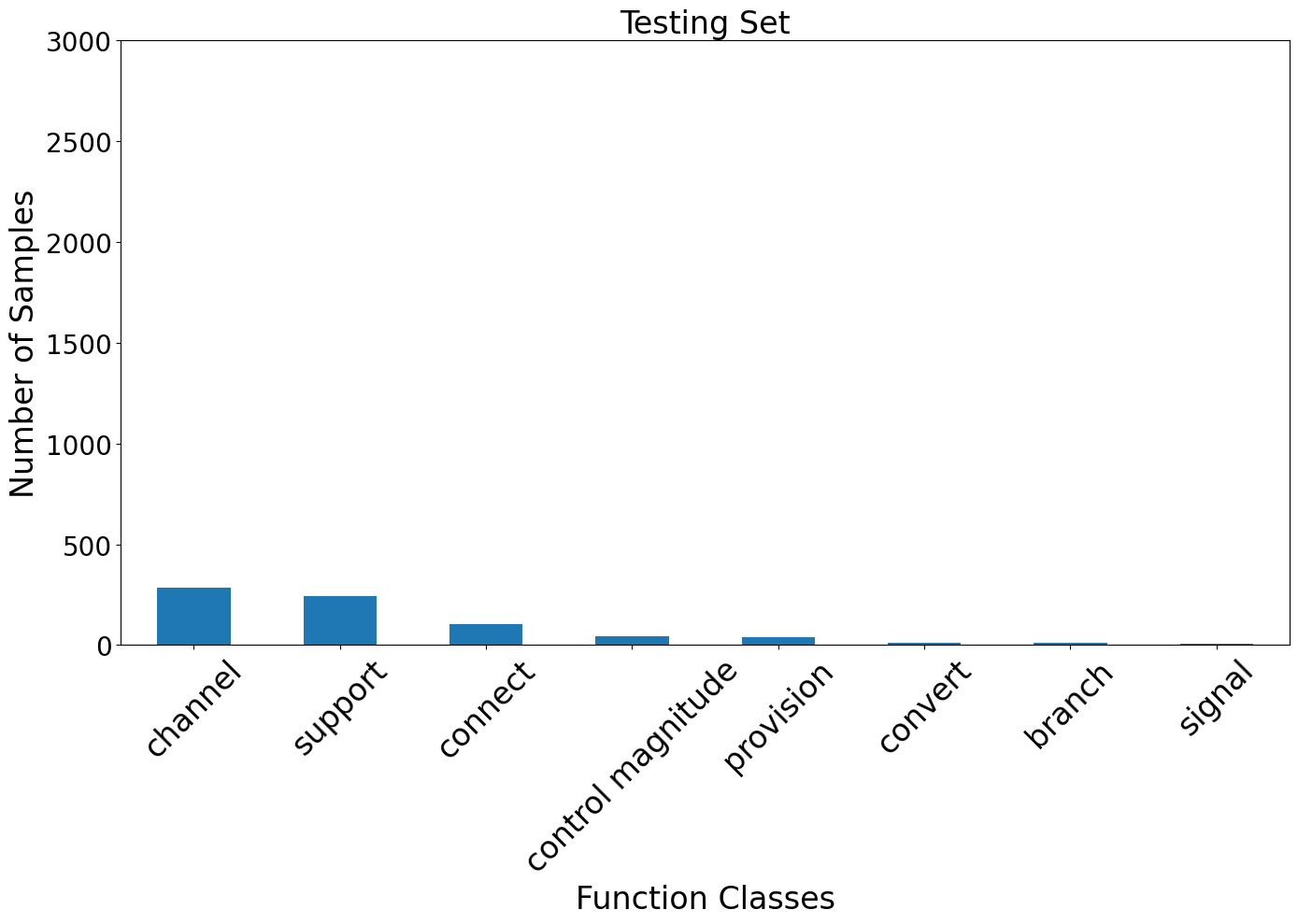}
    \caption{Test set.}
    \label{fig:sub2}
  \end{subfigure}
  
  \caption{The Distribution of Training and Test Set Data for Fine-Tuning Step.}
  \label{Train-Test}
\end{figure*}

We used GPT 3.5 Turbo as the pre-trained LLM for the fine-tuning task. The OSDR, as the function-labeled data, the eight function classes, and the class definitions are the inputs of the LLM. We split the OSDR data to
 $\%90$ and $\%10$ for training and testing, repectively. Figure \ref{Train-Test} indicates the training and test sets of the OSDR data. Additionally, a prompt is used to guide the LLM for the classification task, utilizing the function classes and their definitions: 
\begin{quote}
    Given the following function classes and their definitions, \{function classes: definitions\}, what is the function of a part \{part name\} in the system \{assembly name\}?
\end{quote}

We set different values for hyperparameters during the fine-tuning step. Additionally, to improve the fine-tuning results, we repeated the same experiment with different sample sizes for the training set.

\subsection{Results}
Table \ref{tab:results} illustrates the evaluation results during the fine-tuning step for different hyperparameter values. Accuracy, F1-score, precision, and recall are calculated for both training and test sets to choose the optimum hyperparameter values. 

\begin{table*}[h!]
\centering
\caption{Experimental Results}
\label{tab:results}
\renewcommand{\arraystretch}{2}
\resizebox{\textwidth}{!}{%
\begin{tabular}{|ccc|c|ccc|ccc|ccc|c|ccc|ccc|ccc|}
\hline
\multicolumn{3}{|c|}{\textbf{Hyperparameters}} & \textbf{Train Accuracy} & \multicolumn{3}{c|}{\textbf{Train (Weighted)}} & \multicolumn{3}{c|}{\textbf{Train (Macro)}} & \multicolumn{3}{c|}{\textbf{Train (Micro)}} & \textbf{Test Accuracy} & \multicolumn{3}{c|}{\textbf{Test (Weighted)}} & \multicolumn{3}{c|}{\textbf{Test (Macro)}} & \multicolumn{3}{c|}{\textbf{Test (Micro)}} \\ \cline{1-3} \cline{5-13} \cline{15-23} 
\textbf{Number of Epochs} & \textbf{Batch Size} & \textbf{Learning Rate Multiplier} &  & \textbf{F1 Score} & \textbf{Precision} & \textbf{Recall} & \textbf{F1 Score} & \textbf{Precision} & \textbf{Recall} & \textbf{F1 Score} & \textbf{Precision} & \textbf{Recall} &  & \textbf{F1 Score} & \textbf{Precision} & \textbf{Recall} & \textbf{F1 Score} & \textbf{Precision} & \textbf{Recall} & \textbf{F1 Score} & \textbf{Precision} & \textbf{Recall} \\ \hline
12 & 20 & 20 & 0.64 & 0.64 & 0.64 & 0.64 & 0.31 & 0.31 & 0.31 & 0.63 & 0.63 & 0.63 & 0.42 & 0.43 & 0.42 & 0.42 & 0.29 & 0.30 & 0.28 & 0.4 & 0.4 & 0.4 \\ \hline
20 & 20 & 20 & 0.65 & 0.65 & 0.65 & 0.65 & 0.28 & 0.28 & 0.28 & 0.65 & 0.65 & 0.65 & 0.37 & 0.37 & 0.37 & 0.37 & 0.18 & 0.18 & 0.18 & 0.37 & 0.37 & 0.37 \\ \hline
10 & 48 & 30 & 0.63 & 0.63 & 0.63 & 0.63 & 0.17 & 0.17 & 0.17 & 0.63 & 0.63 & 0.63 & 0.41 & 0.41 & 0.41 & 0.41 & 0.22 & 0.24 & 0.22 & 0.41 & 0.41 & 0.41 \\ \hline
30 & 48 & 30 & 0.66 & 0.66 & 0.66 & 0.66 & 0.51 & 0.51 & 0.51 & 0.66 & 0.66 & 0.66 & 0.37 & 0.37 & 0.37 & 0.37 & 0.20 & 0.21 & 0.20 & 0.37 & 0.37 & 0.37 \\ \hline
8 & 24 & 0.5 & 0.40 & 0.40 & 0.43 & 0.40 & 0.10 & 0.11 & 0.10 & 0.40 & 0.40 & 0.40 & 0.39 & 0.40 & 0.43 & 0.39 & 0.20 & 0.20 & 0.20 & 0.39 & 0.39 & 0.39 \\ \hline
15 & 24 & 20 & 0.64 & 0.64 & 0.65 & 0.64 & 0.27 & 0.27 & 0.27 & 0.64 & 0.64 & 0.64 & 0.41 & 0.41 & 0.41 & 0.41 & 0.25 & 0.25 & 0.24 & 0.41 & 0.41 & 0.41 \\ \hline
12 & 100 & 20 & 0.63 & 0.63 & 0.63 & 0.63 & 0.29 & 0.29 & 0.30 & 0.63 & 0.63 & 0.63 & 0.42 & 0.42 & 0.42 & 0.42 & 0.29 & 0.30 & 0.28 & 0.42 & 0.42 & 0.42 \\ \hline
12 & 20 & 0.6 & 0.42 & 0.43 & 0.45 & 0.42 & 0.15 & 0.14 & 0.16 & 0.42 & 0.42 & 0.42 & 0.35 & 0.36 & 0.38 & 0.35 & 0.16 & 0.16 & 0.18 & 0.35 & 0.35 & 0.35 \\ \hline
12 & 20 & 40 & 0.33 & 0.33 & 0.33 & 0.33 & 0.14 & 0.15 & 0.14 & 0.33 & 0.33 & 0.33 & 0.32 & 0.31 & 0.31 & 0.32 & 0.14 & 0.15 & 0.14 & 0.32 & 0.32 & 0.32 \\ \hline
1 & 20 & 20 & 0.41 & 0.42 & 0.42 & 0.41 & 0.05 & 0.05 & 0.06 & 0.41 & 0.41 & 0.41 & 0.39 & 0.39 & 0.40 & 0.39 & 0.14 & 0.14 & 0.14 & 0.39 & 0.39 & 0.39 \\ \hline
\end{tabular}%
}
\end{table*}

  

\begin{figure*}[h!]
  \centering
  \begin{subfigure}[b]{0.45\textwidth}
    \centering
    \includegraphics[width=\textwidth]{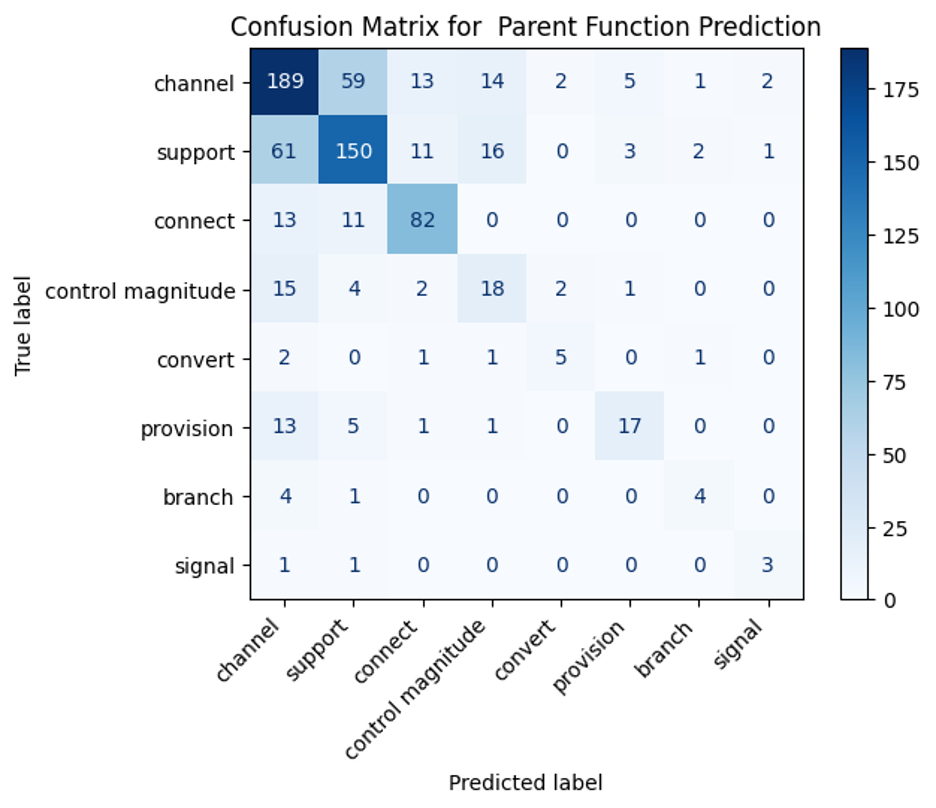}
    \caption{Confusion Matrix for Testing Set for the Fine-Tuned Model with \%90 of Training Samples}
    \label{fig:sub1}
  \end{subfigure}
  \hfill
  \begin{subfigure}[b]{0.45\textwidth}
    \centering
    \includegraphics[width=\textwidth]{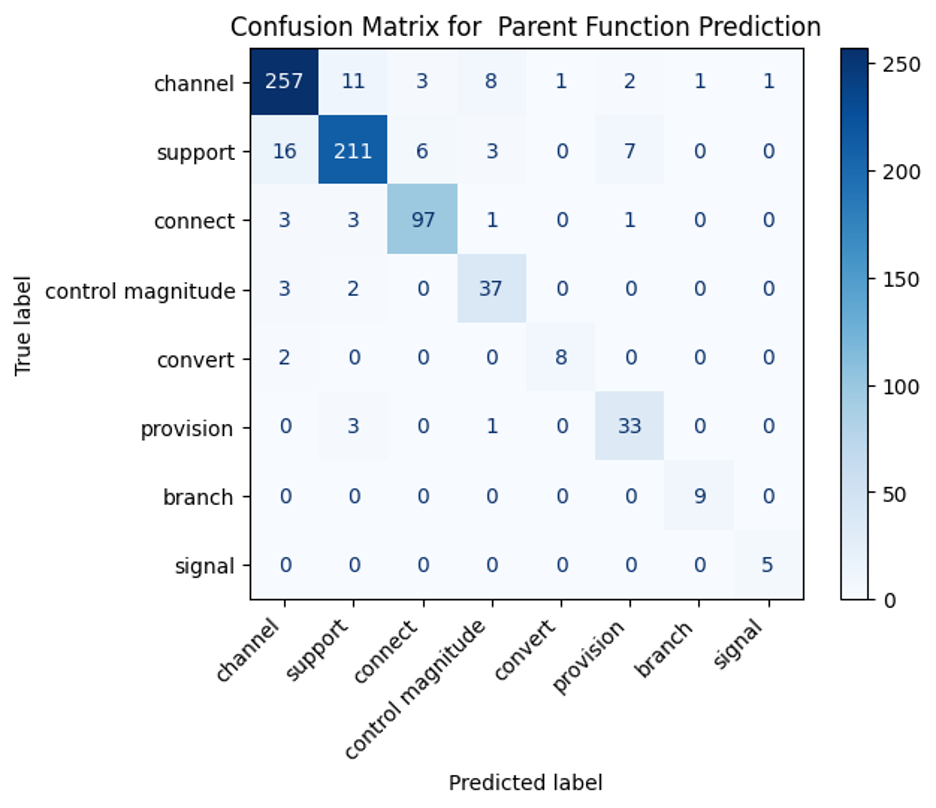}
    \caption{Confusion matrix for Testing Set for the Fine-Tuned Model with \%10 of Training Samples}
    \label{fig:sub2}
  \end{subfigure}
  
  \caption{The confusion matrices for the test set with different training sample sizes.}
  \label{testconf}
\end{figure*}

To improve the fine-tuning results, we trained the model with different training sample sizes for the optimized hyperparameter values. Table \ref{differenttrain} indicates the results of metrics for different training sample sizes. As can be seen, decreasing the data size improves the training and test set prediction results. More specifically, by utilizing only 681 samples of the OSDR data set, we were able to significantly improve the classification results from $\%4$ to $\%90$. Figure \ref{testconf} indicates the confusion matrices for pre-trained LLM with different training set sizes results for the test set.

\begin{table*}[h!]
\centering
\caption{Performance Metrics for Different Training Set Sizes}
\label{differenttrain}
\renewcommand{\arraystretch}{2}
\resizebox{\textwidth}{!}{%
\begin{tabular}{|l|c|ccc|ccc|ccc|c|ccc|ccc|ccc|}
\hline
\textbf{Training Set Size (\%)} & \textbf{Train Accuracy} & \multicolumn{3}{c|}{\textbf{Train (Weighted)}} & \multicolumn{3}{c|}{\textbf{Train (Macro)}} & \multicolumn{3}{c|}{\textbf{Train (Micro)}} & \textbf{Test Accuracy} & \multicolumn{3}{c|}{\textbf{Test (Weighted)}} & \multicolumn{3}{c|}{\textbf{Test (Macro)}} & \multicolumn{3}{c|}{\textbf{Test (Micro)}} \\ \cline{3-11} \cline{13-21} 
 &  & \textbf{F1 Score} & \textbf{Precision} & \textbf{Recall} & \textbf{F1 Score} & \textbf{Precision} & \textbf{Recall} & \textbf{F1 Score} & \textbf{Precision} & \textbf{Recall} &  & \textbf{F1 Score} & \textbf{Precision} & \textbf{Recall} & \textbf{F1 Score} & \textbf{Precision} & \textbf{Recall} & \textbf{F1 Score} & \textbf{Precision} & \textbf{Recall} \\ \hline
10\% (681 samples) & 0.93 & 0.93 & 0.94 & 0.93 & 0.82 & 0.80 & 0.84 & 0.93 & 0.93 & 0.93 & 0.89 & 0.89 & 0.90 & 0.89 & 0.64 & 0.63 & 0.66 & 0.89 & 0.89 & 0.89 \\ \hline
30\% (2043 samples)  & 0.83 & 0.83 & 0.84 & 0.83 & 0.64 & 0.64 & 0.64 & 0.83 & 0.83 & 0.83 & 0.84 & 0.84 & 0.85 & 0.84 & 0.84 & 0.85 & 0.83 & 0.84 & 0.84 & 0.84 \\ \hline
40\% (2724 samples)  & 0.78 & 0.78 & 0.78 & 0.78 & 0.66 & 0.64 & 0.67 & 0.78 & 0.78 & 0.78 & 0.79 & 0.79 & 0.79 & 0.79 & 0.56 & 0.56 & 0.56 & 0.79 & 0.79 & 0.79 \\ \hline
50\% (3406 samples)  & 0.75 & 0.75 & 0.75 & 0.75 & 0.39 & 0.39 & 0.40 & 0.75 & 0.75 & 0.75 & 0.75 & 0.75 & 0.75 & 0.75 & 0.72 & 0.71 & 0.72 & 0.75 & 0.75 & 0.75 \\ \hline
75\% (5109 samples)  & 0.68 & 0.69 & 0.69 & 0.68 & 0.31 & 0.30 & 0.31 & 0.68 & 0.68 & 0.68 & 0.68 & 0.68 & 0.69 & 0.68 & 0.55 & 0.56 & 0.55 & 0.68 & 0.68 & 0.68 \\ \hline
90\% (6130 samples)  & 0.64 & 0.64 & 0.64 & 0.64 & 0.34 & 0.34 & 0.34 & 0.64 & 0.64 & 0.64 & 0.63 & 0.63 & 0.64 & 0.63 & 0.56 & 0.57 & 0.56 & 0.63 & 0.63 & 0.63 \\ \hline
\end{tabular}%
}
\end{table*}

\begin{figure}[h!]
    \centering
    \includegraphics[width=\columnwidth, height=10\textheight, keepaspectratio]{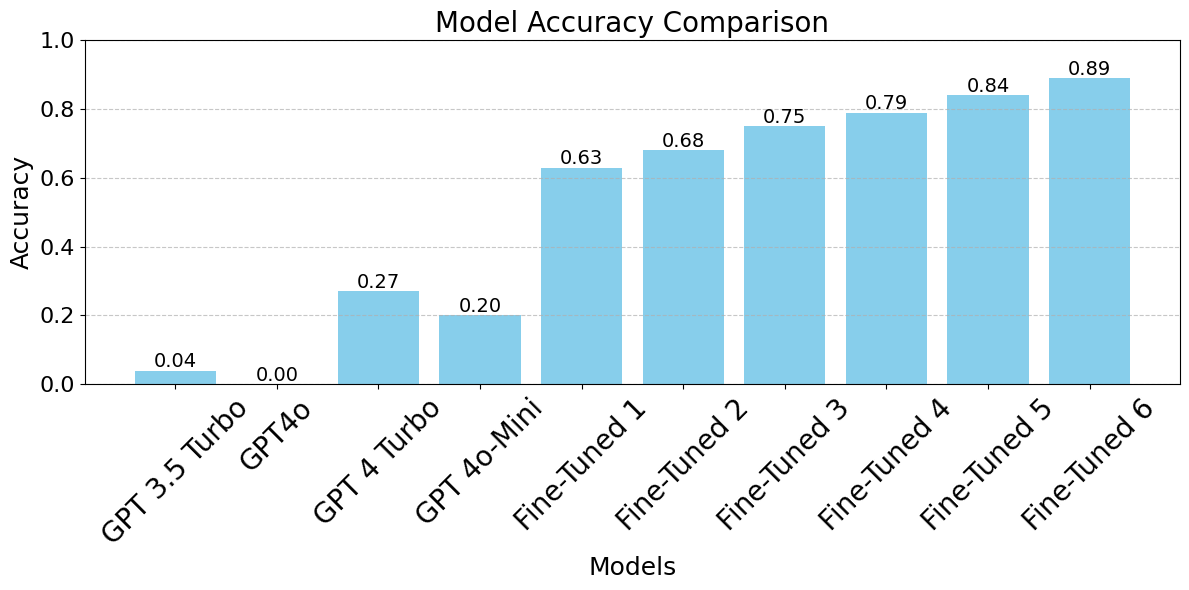}
    \caption{THE COMPARISON OF ACCURACY FOR PRE-TRAINED GPT MODELS AND FINE-TUNED MODELS}
    \label{Com}
\end{figure}

 As a comparative study, we compared the prediction accuracy of the fine-tuned models, containing different training sample sizes, with that of the latest pre-trained GPT models, as can be seen in Figure \ref{Com}. Table \ref{tab:model_comparison} indicates the detailed results of the pre-trained GPTs. The fine-tuned LLM with different training sample sizes all outperformed the pre-trained GPT models, indicating the importance of fine-tuning LLMs for classifying mechanical assembly parts' functions.
 
\begin{table*}[h!]
\centering
\caption{Performance Evaluation of Pre-trained GPT Models on Test Data}
\label{tab:model_comparison}
\renewcommand{\arraystretch}{1}
\resizebox{\textwidth}{!}{%
\begin{tiny}  
\begin{tabular}{|l|c|ccc|ccc|ccc|}
\hline
\textbf{Model} & \textbf{Accuracy} & \multicolumn{3}{c|}{\textbf{Weighted}} & \multicolumn{3}{c|}{\textbf{Macro}} & \multicolumn{3}{c|}{\textbf{Micro}} \\ \cline{3-11} 
  &  & \textbf{F1 Score} & \textbf{Precision} & \textbf{Recall} & \textbf{F1 Score} & \textbf{Precision} & \textbf{Recall} & \textbf{F1 Score} & \textbf{Precision} & \textbf{Recall} \\ \hline
GPT-3.5 Turbo & 0.04 & 0.06 & 0.15 & 0.04 & 0.02 & 0.04 & 0.01 & 0.04 & 0.04 & 0.04 \\ \hline
GPT-4 Turbo   & 0.27 & 0.28 & 0.39 & 0.27 & 0.06 & 0.07 & 0.07 & 0.27 & 0.27 & 0.27 \\ \hline
GPT-4o Mini   & 0.20 & 0.22 & 0.43 & 0.20 & 0.07 & 0.09 & 0.09 & 0.20 & 0.20 & 0.20 \\ \hline
\end{tabular}%
\end{tiny}  
}
\end{table*}

\begin{figure}[h!]
    \centering
    \includegraphics[width=\columnwidth]{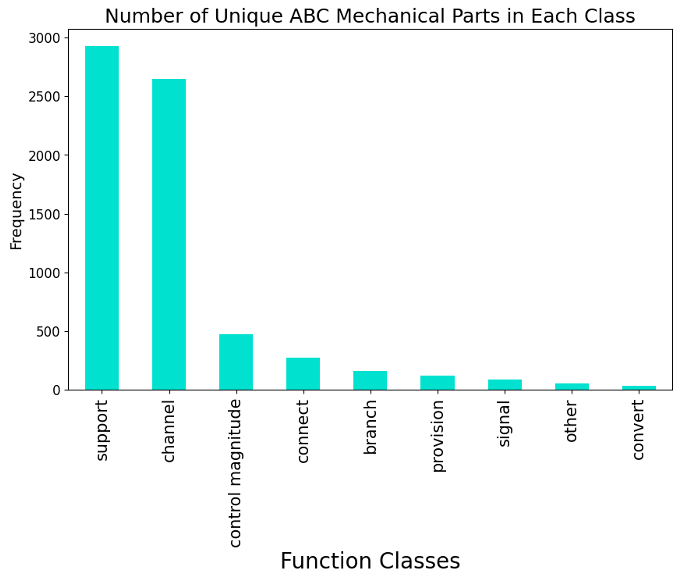}
    \caption{Generated Synthesized Functions of ABC Mechanical Assembly Parts}
    \label{Classification}
\end{figure}

 We applied the fine-tuned model with the highest accuracy to the matched ABC dataset for function classification. Figure \ref{Classification} indicates the results of function annotation for the ABC data set. As shown, $\%99$ of the classified labels fall within the pre-defined function classes, with only $\%1$ of the samples being misclassified. Additionally, 'channel' and 'support' are the most frequently classified labels, consistent with the function class distribution in the OSDR data. Table \ref{tab:sample_table} indicates samples of newly generated function-labeled ABC data.

\begin{table*}[h!]
\centering
\caption{Samples of Generated Synthesized Function Labels for the ABC MECHANICAL PARTS}
\begin{tiny}  
\resizebox{\textwidth}{!}{%
\begin{tabular}{|c|c|c|c|c|c|c|}
\hline
\textbf{Chunk} & \textbf{Assembly id} & \textbf{Assembly name} & \textbf{Part id} & \textbf{Part name} & \textbf{Function} & \textbf{Image} \\ \hline
0 & 390 & Tablet Stand & 39 & Washer & Support & \includegraphics[width=1.5cm]{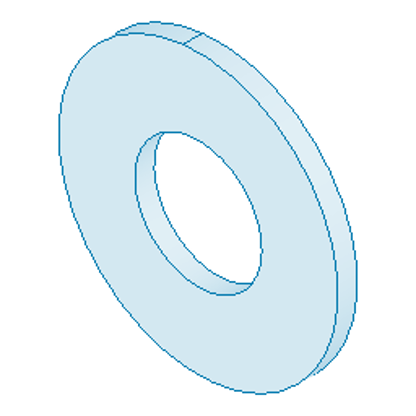} \\ \hline
0 & 253 & Independent Front Suspension & 18 & Spindle & Channel & \includegraphics[width=1.5cm]{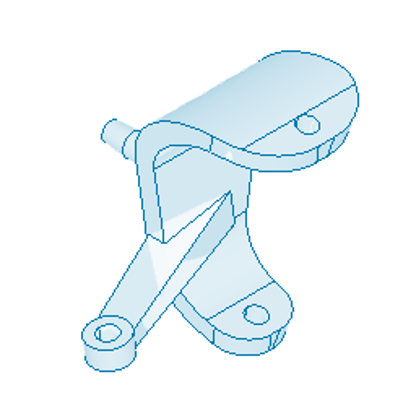} \\ \hline
\end{tabular}%
}
\end{tiny}
\label{tab:sample_table}
\end{table*}

\section{DISCUSSION}
DA of LLMs has significantly improved the function classification results of mechanical assembly parts as compared to general-purpose LLMs, as shown in Figure \ref{Com}. Moreover, LLM's unique capabilities allowed the incorporation of function definitions as an extra input for both DA and classification steps. This integration, which was not feasible by traditional ML methods, enabled the model to better understand the context and relationships
between component descriptions and their associated functions.

Despite the noticeable capability of LLMs DA in generating synthesized function-labeled conceptual design data, several limitations and challenges were observed. First, the prediction accuracy during fine-tuning improved when the number of training samples decreased. One possible reason could be that LLMs, specifically GPTs, for this specific domain, struggle to handle larger datasets, resulting in overfitting.

Second, the OSDR data used for the DA step is imbalanced, with certain function classes, such as channel and support, being heavily overrepresented, as shown in Figure \ref{Train-Test}. The
model’s strong performance in these classes is likely a reflection of their dominance in the training data. In contrast, the accuracy for less common function
classes was significantly lower, revealing a bias towards more frequently occurring function classes as shown in Figure \ref{Classification}. 

Third, the DA prediction results heavily depend on the pre-defined function labels, limiting the domain-adapted classifier's ability to distinguish other possible function classes. As shown in Figure \ref{Classification} $\%99$ of the classified labels fit within the
existing classes.

\section{CONCLUDING REMARKS AND FUTURE WORKS}

The study developed a supervised DA framework for the function classification of mechanical assembly parts. The methodology aimed to fine-tune a foundational LLM model, GPT 3.5 Turbo, utilizing the OSDR data. The DA approach successfully generated synthesized functional labels for the ABC data set to address the lack of function data in conceptual design.

Future work will focus on improving the model performance, utilizing reasoning-enhanced and chain-of-thoughts prompting approaches. Additionally, DA will be applied to Vision-Language models utilizing both textual and vision design data to predict the functions of mechanical assembly components.




\nocite{*}



\end{document}